\documentclass[10pt,twocolumn,letterpaper]{article}

\usepackage[pagenumbers]{cvpr} 
\usepackage{algorithmic}
\usepackage{algorithm}
\usepackage{makecell}

\definecolor{cvprblue}{rgb}{0.21,0.49,0.74}
\usepackage[pagebackref,breaklinks,colorlinks,allcolors=cvprblue]{hyperref}

\def\paperID{} 
\def\confName{}
\def\confYear{}

\title{Consistent Image Layout Editing with Diffusion Models}

\author{Tao Xia\\
Beijing Institute of Technology\\
China\\
\and
Yudi Zhang\\
Beijing Institute of Technology\\\
China\\
\and
Lei Zhang\\
Beijing Institute of Technology\\\
China\\
}

\begin{document}
\twocolumn[{%
\renewcommand\twocolumn[1][]{#1}%
\maketitle
\begin{center}
    \centering
    \includegraphics[width=1.0\linewidth]{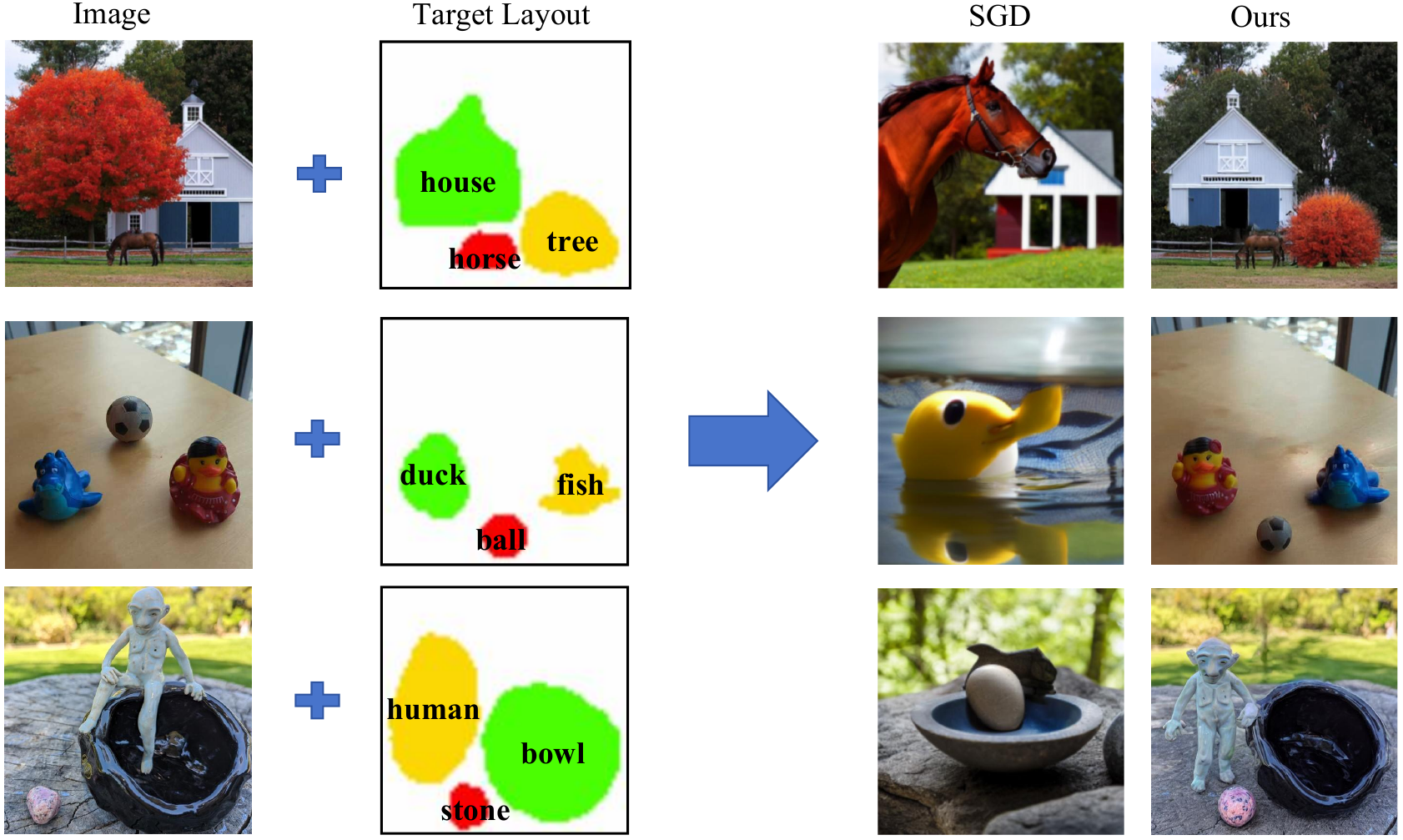}
    \captionof{figure}{\textbf{Examples of layout editing for a single real image.} Given a single real image, our method can be used to transform its layout and preserve consistent visual appearance compared to self-guidance-diffusion(SGD \cite{epstein2023selfguidance}).}
\end{center}%
}]
\begin{abstract}
Despite the great success of large-scale text-to-image diffusion models in image generation and image editing, existing methods still struggle to edit the layout of real images. Although a few works have been proposed to tackle this problem, they either fail to adjust the layout of images, or have difficulty in preserving visual appearance of objects after the layout adjustment. To bridge this gap, this paper proposes a novel image layout editing method that can not only re-arrange a real image to a specified layout, but also can ensure the visual appearance of the objects consistent with their appearance before editing. 
Concretely, the proposed method consists of two key components. Firstly,  a multi-concept learning scheme is used to learn the concepts of different objects from a single image, which is crucial for keeping visual consistency in the layout editing. 
Secondly, it leverages the semantic consistency within intermediate features of diffusion models to project the appearance information of objects to the desired regions directly. 
Besides, a novel initialization noise design is adopted to facilitate
the process of re-arranging the layout. Extensive experiments demonstrate that the proposed method outperforms previous works in both  layout alignment and visual consistency for the task of image layout editing.\end{abstract}
\section{Introduction}
\label{sec:intro}

Image generation based on diffusion models belongs to central topics in the field of artificial intelligence generative content (AIGC) \cite{rombach2022high, ho2020denoising, ramesh2021zero, song2020denoising, podell2023sdxl}. It dedicates to generate high-quality images that meet specific semantics and layout arrangement by taking text, semantic maps, or other conditions as inputs. Nevertheless, re-arranging objects in an existing image is even more valuable and more difficult. There are two main challenges in image layout editing task. One is transforming the original image according to the target layout specified by users, the other one is keeping the appearance of the objects unchanged after re-arranging the layout.

Some of research has focused on the layout control methods in the process of image generation \cite{xie2023boxdiff, couairon2023zero, chen2024training}. Despite of their great success, however, these methods can only control the layout for generated images and hardly to be adapted to an existing image. Continuous layout editing with diffusion (CLED) \cite{zhang2023continuous} proposes the first framework for editing the layout of real images. It learns the concepts of multiple objects distributed in a single image and then achieves the layout adjustment by optimizing the latent of diffusion models. However, it struggles to maintain the consistency of the color, shape, and texture of the objects after the layout editing. 
\begin{figure}[t]
  \centering
   \includegraphics[width=1.0\linewidth]{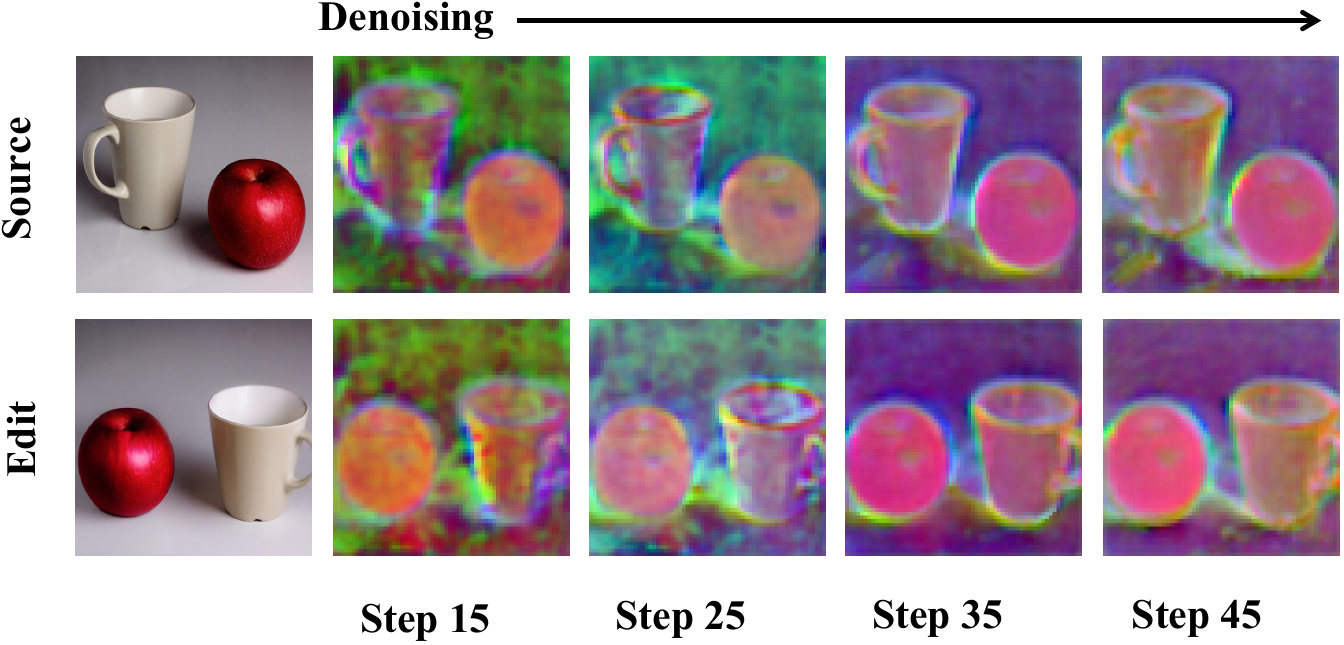}
   \caption{\textbf{Semantic consistency in diffusion feature space.}
   The first column shows the original image and the layout editing result by our method, and the following columns show the principal component analysis (PCA) \cite{pearson1901closestfit} of intermediate diffusion features. The similar semantics share similar colors. It shows that semantic consistency in the RGB space can be extended into diffusion feature space across the whole denoising process}
   \label{fig:motivation}
\end{figure}

It is observed that the image layout editing is a typical structural editing task. It only updates the structure of the image while keeping its semantics unchanged. In other words, there is semantic consistency between the original and edited images in the RGB space. Additionally, Plug-and-Play \cite{tumanyan2023plug} and TokenFlow \cite{geyer2023tokenflowconsistentdiffusionfeatures} have proven that the semantic consistency can be extended to the intermediate feature space of the diffusion models as shown in \cref{fig:motivation}. This inspires us to utilize this principle to project the appearance information of the objects from the original image to the target areas in the edited results, which should be able to improve the fidelity of the objects after the layout adjustment. Therefore, we propose a two-stage image layout editing framework that can not only re-arrange the layout of an image properly, but also can ensure the visual consistency of the objects. Specifically, we explore a multi-concept learning method suitable for layout editing task, enabling the extraction of multiple concepts from a single image. Then, we leverage the semantic consistency within the diffusion feature space and incorporate layout priors to develop an effective appearance projection technique, which is important to keep the appearance consistency after the layout adjustment. Besides, it is observed that the initial noise has a significant impact on the image layout. So by carefully designing the initial noise, we can facilitate the process of layout adjustment and enhance the fidelity of the objects.

The main contribution of this paper is a novel two-stage image layout editing method by  leveraging the semantic consistency of the diffusion models. It can re-arrange the layout of an existing image and ensure consistency in appearance of the objects. Besides, a layout-friendly initialization noise (LFIN) strategy is designed to facilitate the process of layout adjustment.
To evaluate the effectiveness of the proposed method, we also collect the first public dataset designed for image layout editing, namely Layout-Bench. Experimental evidence shows the effectiveness of the proposed method in the image layout editing task.

\section{Related work}
\label{sec:intro}

In recent years, text-to-image diffusion models, including stable diffusion (SD) \cite{rombach2022high} and SDXL \cite{podell2023sdxl}, have made significant progress. It can be attributed to large-scale training datasets like Laion-400M \cite{schuhmann2021laion} and Conceptual-12M \cite{changpinyo2021conceptual}. However, text-based image generation models often fail to generate images with desired layouts, which requires more conditions about layout arrangement to control the generated images.

\noindent\textbf{Layout Control in Image Generation. } It is challenging to precisely control the relative positions of several objects and the layout of generated images using pre-trained Text2Image diffusion models \cite{rombach2022high, podell2023sdxl}. 
To address this limitation, several methods for controlling the layout have been proposed \cite{li2023gligen, yang2022reco, zhang2023adding, couairon2023zero, xie2023boxdiff}, which can be divided into three categories. 
The first category requires model fine-tuning \cite{li2023gligen, yang2022reco, zhang2023adding}. 
For instance, GLIGEN \cite{li2023gligen} and ReCo \cite{yang2022reco} employ gated self-attention layers and additional region tokens respectively, to fine-tune diffusion models. 
ControlNet \cite{zhang2023adding} inserts additional conditions, such as the semantic maps for layout control, by utilizing a trainable copy of the original U-Net \cite{ho2020denoising} model. The conditional copy and the original model are fused in intermediate layers to generate a conditioned output.
These methods require adding additional modules to the pre-trained diffusion models and necessitate paired data (e.g., images and corresponding semantic segmentation maps) to fine-tune the diffusion models and the added modules. Moreover, their capabilities are constrained by the training data.
The second category explores layout control in a training-free way \cite{bar-tal2023multidiffusion, liu2023cones, singh2023high}. The representative work is MultiDiffusion \cite{bar-tal2023multidiffusion}, which denoises different crops of each object locally and then fuses the results globally for each denoising step.
Cones \cite{liu2023cones} and HFG \cite{singh2023high} propose to directly intervene in the cross-attention. The third category involves methods based on latent optimization, such as Dragdiffusion \cite{dragdiffusion}, Directed Diffusion \cite{ma2023directeddiffusiondirectcontrol}, ZestGuide \cite{couairon2023zero}, and BoxDiff \cite{xie2023boxdiff}. They propose to build layout loss based on the cross-attention map, then optimize latent in diffusion models so that the objects can arrive at the specified positions. Layout Guidance \cite{chen2024training} further provides a detailed analysis of the rationality for this method.

\noindent\textbf{Image Editing Based on Diffusion Models.}
Most aforementioned methods lack the ability to modify an already generated image, or even the ability to edit real images.
Image editing based on diffusion models aims to modify specific areas of an image according to user instructions, such as adding, deleting, or replacing objects, or adjusting attributes like color, pose and material, while preserving the remaining content of the image.
However, most research focuses on editing images in terms of semantics \cite{DiffusionClip, imagic, instructp2p} or style \cite{tumanyan2023plug, hertz2022prompt, matsunaga2022finegrained}, fewer studies pay attention to reconstructing the layout of an existing image.
CLED \cite{zhang2023continuous} proposes the first framework for image layout editing, but its output often tends to present distorted appearance of the objects.
Inspired by the concept of layers in the design field, DesignEdit \cite{jia2024designeditmultilayeredlatentdecomposition} proposes a two-stage multi-layer editing framework based on latent diffusion models (LDMs) \cite{podell2023sdxl}. However, it may face challenges in ensuring that the lighting and shadows in the generated image adhere to real-world physical laws, resulting in a lack of realism in the edited outputs and limitations in handling occlusion scenarios.

\noindent\textbf{Semantic Consistency in Diffusion Models.}
Plug-and-Play \cite{tumanyan2023plug} finds that diffusion models can capture the fine-grained semantic information and reveals  the semantic consistency existing in intermediate features of diffusion models.
DreamMatcher \cite{DreamMatcher} proposes to utilize the semantic correspondence to project the appearance information from the reference image into the target structure. It can ensure consistency of the objects identity in personalized images.
Furthermore, TokenFlow \cite{geyer2023tokenflowconsistentdiffusionfeatures} finds that the internal representation of the video in the diffusion model exhibits similar properties. That is, the level of redundancy and temporal consistency of the frames in the RGB space and in the diffusion feature space are tightly correlated.

\section{Method}
\label{sec:method}
\begin{figure*}[]
  \centering
   \includegraphics[width=1.0\linewidth]{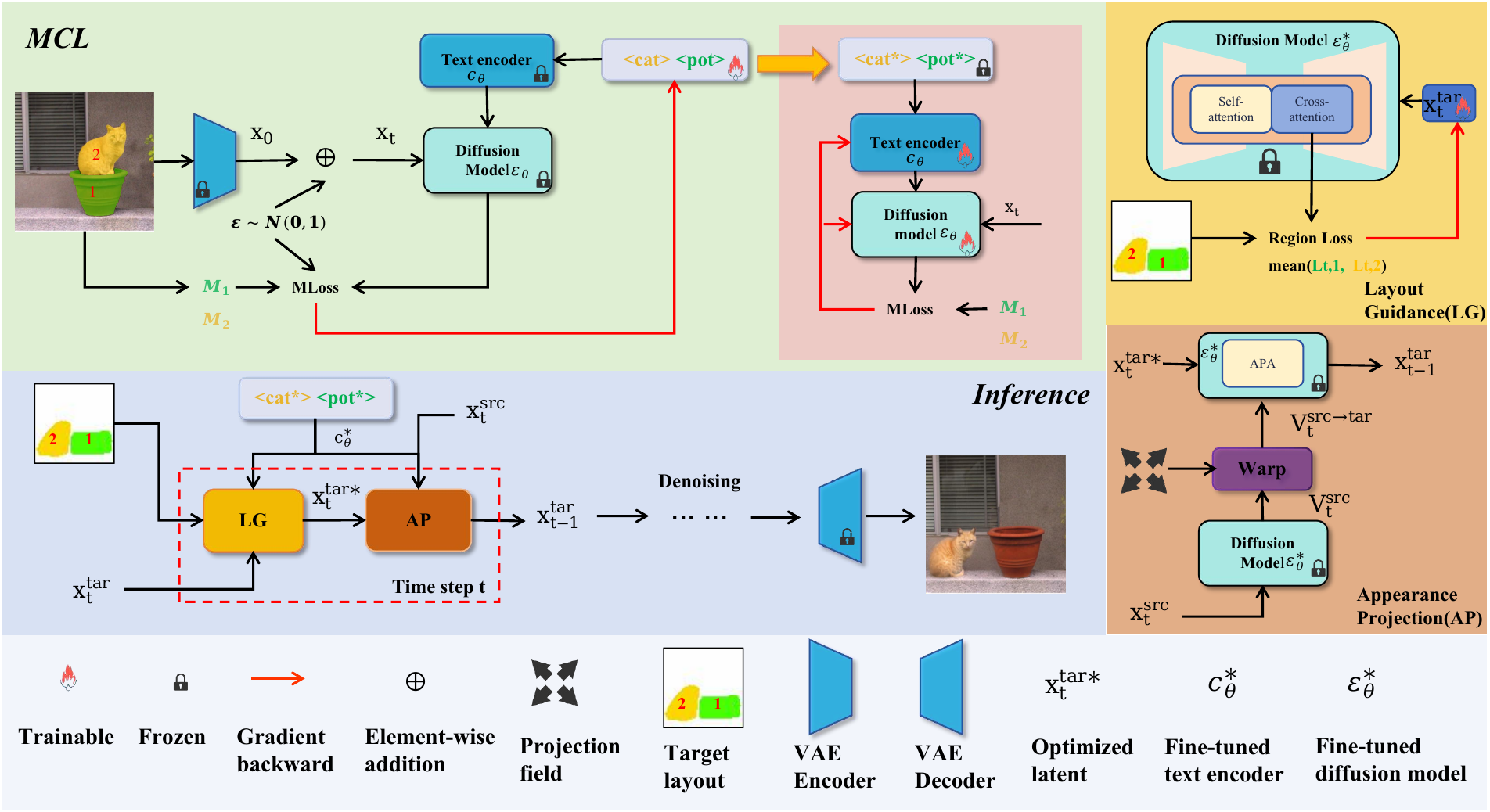}

   \caption{\textbf{Method overview.}}
   \label{fig:pipeline}
\end{figure*}
Given an original image $I^{src}$ and its layout $L^{src}$, which can be provided by the users or automatically obtained by using a segmentation model, such as the segment anything model (SAM) \cite{kirillov2023segment}. Additionally, the user needs to provide a target layout $L^{tar}$, and our goal is to obtain the layout editing result $I^{tar}$. The framework of our method is illustrated in \cref{fig:pipeline}, where we firstly apply multi-concept learning (MCL) to extract multiple concepts from 
$I^{src}$, and then we perform layout guidance followed by appearance projection at each denoising step. Our method is based on the state-of-the-art text-to-image generation model of Stable Diffusion \cite{rombach2022high}.

\subsection{Multi-Concept Learning in a Single Image}
\label{subsec: mcl}
\begin{figure}[h]
  \centering
   \includegraphics[width=1.0\linewidth]{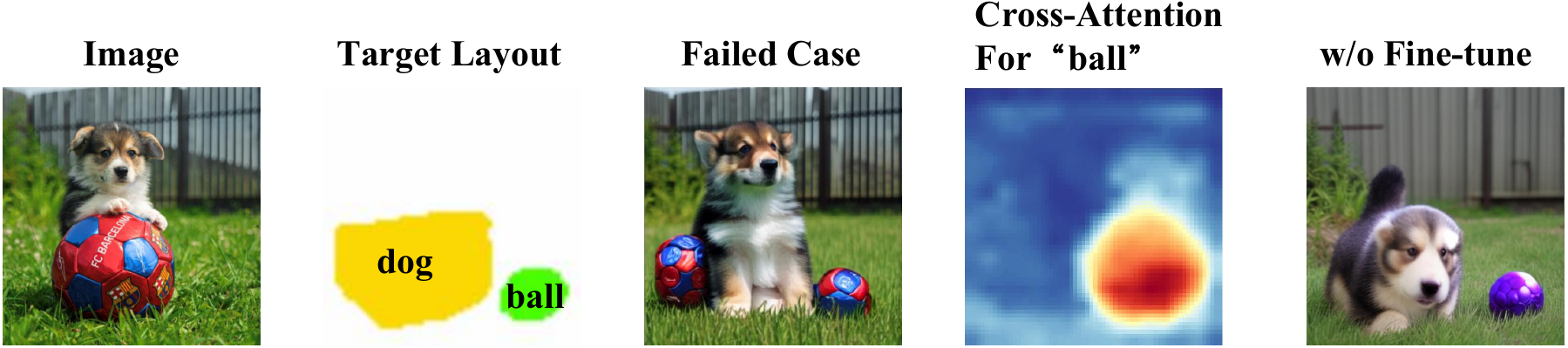}

   \caption{\textbf{Failed case and analysis.} The attention region edited by CLED \cite{zhang2023continuous} is spread across the entire image rather than focusing on the object. The right column shows the editing results by removing fine-tuning stage.}
   \label{fig:badecase-cled}
\end{figure}

To preserve the visual appearance after editing, we initially conduct multi-concept learning on several objects within a single image.
However, learning multiple concepts from a single image is inherently more challenging than learning a single concept from a set of images \cite{hu2021loralowrankadaptationlarge, kumari2023CD, ruiz2023dreamboothfinetuningtexttoimage}, since the available information reduces while the number of concepts to be learned increases.
CLED \cite{zhang2023continuous} employs a two-stage training scheme, wherein it first optimizes only the textual embeddings of the target concepts with masked diffusion loss (MLoss), followed by a fine-tuning stage by using Custom-Diffusion (CD) \cite{kumari2023CD}.
However, our experiments reveal that there are two main weaknesses: the duplication of objects and the failure of layout re-arrangement as shown in \cref{fig:badecase-cled}.

By visualizing the cross-attention map corresponding to the failure objects, we find that the attention regions are often spread across the entire image rather than focusing on the objects (see the 4-th column in \cref{fig:badecase-cled}.). This is a typical phenomenon of overfitting \cite{tewel2023keylocked}.
Furthermore, an ablation study on the fine-tuning stage indicates that it can alleviate the issue of transformation failure by removing this stage, while at the cost of noticeable changes in the appearance of the objects (see the 5-th column in \cref{fig:badecase-cled}.). Additionally, it is observed that the masked diffusion loss plays an important role in the fine-tuning stage.
Therefore, we propose a multi-concept learning scheme suitable for the task of layout editing. The workflow is depicted in \cref{fig:pipeline}. Similar to CLED \cite{zhang2023continuous}, our method also consists of two stages. The first stage employs the same processing as CLED \cite{zhang2023continuous}, only optimizing textual embeddings of target concepts to learn an unique token embedding for each object. 
Differently, we continue to apply the masked diffusion loss and fine-tune all layers of the diffusion models in the second stage, not just the key and value weights of cross-attention layers suggested by CD \cite{kumari2023CD}. Because our experiments show that only fine-tuning these layers, with the masked diffusion loss, would significantly compromise the fidelity of the edited objects (see the experiments in \cref{sec:abalation}). The formula for the masked diffusion loss is defined as follows:
\begin{equation}
    L_{masked} = E_{x,i,\epsilon \sim N(0,1),t}[||\epsilon \odot M_i-\epsilon_\theta(x_t,t,p_i) \odot M_i||]
  \label{eq:masked diffusion loss}
\end{equation}
where $M_i$ is the mask of the \textit{i}-th object (\textit{i}=1, ..., \textit{N} is the index of the \textit{N} objects), and $\epsilon_\theta(x_t,t,p_i)$ denotes the output of the denoising U-Net \cite{song2020denoising}, given the noisy latent $x_t$ with the time step $t$ and text prompt $p_i$ as inputs.
\subsection{Layout Guidance}
To achieve layout control, there are some methods based on latent optimization, such as Layout Guidance \cite{chen2024training}, BoxDiff \cite{xie2023boxdiff} and Directed Diffusion \cite{ma2023directeddiffusiondirectcontrol} that utilize cross-attention map to construct a layout loss and optimize latent through back propagation, guiding the diffusion process towards the desired layout. Inspired by previous works, we construct a region loss based on the cross-attention as shown in \cref{eq:region loss}. At each denoising step, the region loss for every object is calculated separately and then averaged. Concretely, the loss function is defined by: 
\begin{equation}
    L (A_{i,t},M_i,i)_{i,t} = 1-\frac{\sum_{u \in M_i}A_{i,t,u}}{\sum_u A_{i,t,u}}
  \label{eq:region loss}
\end{equation}
\begin{equation}
    L_{total,t} = \frac{1}{N} \displaystyle \sum^{N}_{i=1}L_{i,t}
  \label{eq:mean region loss}
\end{equation}
where $A_{i,t} = \frac{1}{L} \sum^{L}_{l=1}A_{i,l,t}$ represents the averaged cross attention of all layers at the time step t. The gradient of the region loss is computed to update the latent $x_t$ via back propagation as follows:
\begin{equation}
    x_t \leftarrow x_t - \sigma^2_t\eta\nabla_{x_t}L_{total,t}
  \label{eq:latent-optmization}
\end{equation}
where $\eta>0$ is a scale factor controlling the strength of the guidance and $\sigma_t = \sqrt{\frac{(1 - \alpha_t)}{\alpha_t}}$, $\alpha_t$ is a function of t that decreases monotonically \cite{nie2024blessingrandomnesssdebeats}.
Since the image layout is often formed in the early stage of denoising \cite{hertz2022prompt}, we only apply region loss to optimize latent in the early steps.
\subsection{Appearance Projection}
Layout control based on latent optimization always leads to the distortion of the object appearance after editing \cite{jia2024designeditmultilayeredlatentdecomposition}. Therefore, we believe that it is necessary to perform appearance compensation after the transformation. Inspired by Plug-and-Play \cite{tumanyan2023plug}, TokenFlow \cite{geyer2023tokenflowconsistentdiffusionfeatures} and DreamMatcher \cite{DreamMatcher}, we consider that we can fully utilize the semantic consistency in the intermediate features of the diffusion models to establish a projection field, and project the appearance information from the original image to the specified area, which should be able to repair the appearance of objects.

\noindent\textbf{Unconditional Appearance Projection.} Plug-and-Play \cite{tumanyan2023plug} mentions that the intermediate features of the diffusion models contain fine-grained semantic information, which inspires us to use the intermediate features of the diffusion models to establish a projection field to project the appearance information of targets in the original image to the specified locations. To achieve this goal, we adopt a source-target dual-branch structure similar to that used in P2P \cite{hertz2022prompt} and Masactl \cite{cao2023masactrl}.
We invert $I^{src}$ to $X^{src}_{T}$ by DDIM inversion \cite{song2020denoising} and save the trace $\{X^{src}_t, t=1, 2, ..., T\}$, while $I^{tar}$ is generated from a random Gaussian noise $X^{tar}_T$ guided by the prompt $p$. At each time step, the self-attention modules from the source branch projects image features into $Q^{src}_t$, $K^{src}_t$, and $V^{src}_t$, while the target branch produces $Q^{tar}_t$, $K^{tar}_t$, and $V^{tar}_t$. Let $\epsilon_{\theta,l}(x_t,t,p)$ denotes the output of the $l$-th decoder layer of the denoising U-Net \cite{ho2020denoising} at the time step $t$. Then, we extract the feature descriptor $f_{l,t}$ as follows:
\begin{equation}
    f_{l,t} = \epsilon_{\theta,l}(x_t,t,p)
  \label{eq:unet}
\end{equation}

\noindent We subsequently extract the intermediate features of the original image and the target image in the $l$-th layer of the U-Net, denoted as $f^{src}_{l,t}$ and $f^{tar}_{l,t}$ respectively, then perform principal component analysis (PCA) \cite{pearson1901closestfit} to reduce the dimensionality.
We build the similarity matrix by calculating the pairwise cosine similarity between feature descriptors extracted from the source and target branches. It is formulated as:
\begin{equation}
    Sim_t\left(i,j\right) = \frac{f^{src}_t\left(i\right) \cdot f^{tar}_t\left(j\right)}{||f^{src}_t\left(i\right)|||f^{tar}_t\left(j\right)|||}
  \label{eq:Sim}
\end{equation}
where $f^{src}_{t}$ and $f^{tar}_{t}$ represent the concatenated features of multiple layers for $f^{src}_{l,t}$ and $f^{tar}_{l,t}$, and $||\cdot||$ represents L2 normalization. After obtaining the similarity matrix $Sim_t \in R^{hw \times hw}$, we perform $argmax$ operation to obtain the projection field $P \in R^{h \times w}$, where h,w refer to the height and width of latent.
Afterward, we can project the appearance information of the objects in the original image to the edited results according to $P$ as defined in \cref{eq:warp}. We only project $V^{src}_t$ features, because query and key are more related to the structure of the image as concluded in DreamMatcher \cite{DreamMatcher} and PerFusion \cite{tewel2023keylocked}.
\begin{equation}
    V^{src\rightarrow tar}_t = W\left(V^{src}_t,P\right)
  \label{eq:warp}
\end{equation}
$W$ represents the warping operation \cite{truong2021warpconsistency}. Unconditional appearance projection (UAP) then implants appearance projection by replacing $V^{tar}_t$ with $V^{src\rightarrow tar}_t$ in the self-attention modules of target branch, which is called appearance projection self-attention (APA). It is formulated as:

\begin{small}
\begin{equation}
    APA(Q^{tar}_t,K^{tar}_t,V^{src\rightarrow tar}_t) = Softmax(\frac{Q^{tar}_t\left(K^{tar}_t\right)^\mathrm{T}}{\sqrt{d}})V^{src\rightarrow tar}_t
  \label{eq:self-attention}
\end{equation}
\end{small}


\noindent where $d$ represents the channel dimensions of $V^{src\rightarrow tar}_t$.

\begin{figure}[htbp]
  \centering
   \includegraphics[width=1.0\linewidth]{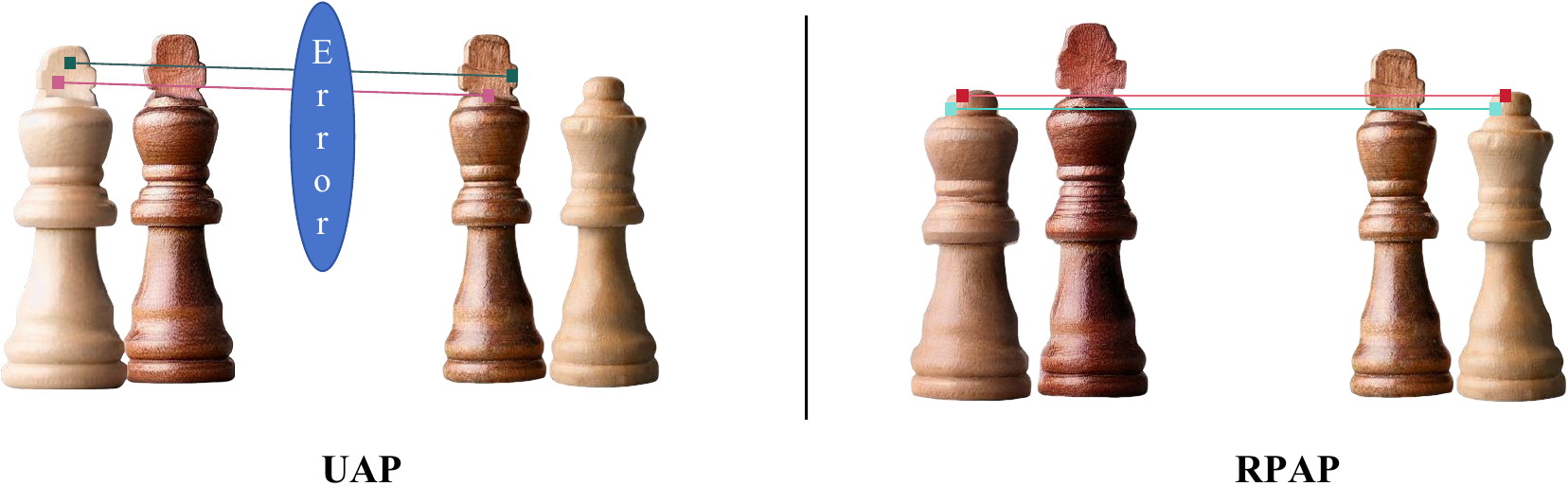}

   \caption{\textbf{Mismatched case. }
   Left: Mismatched points by UAP. Right: Corrected matching points by RPAP}
   \label{fig:mismatch}
\end{figure}

\noindent\textbf{Region Prior Appearance Projection.} Through UAP, we can successfully establish a projection field and implement appearance projection. However, we find that unconditional feature matching often leads to wrong matched points, which affects the editing results as shown in \cref{fig:mismatch}.

\begin{figure}[h]
  \centering
   \includegraphics[width=1.0\linewidth]{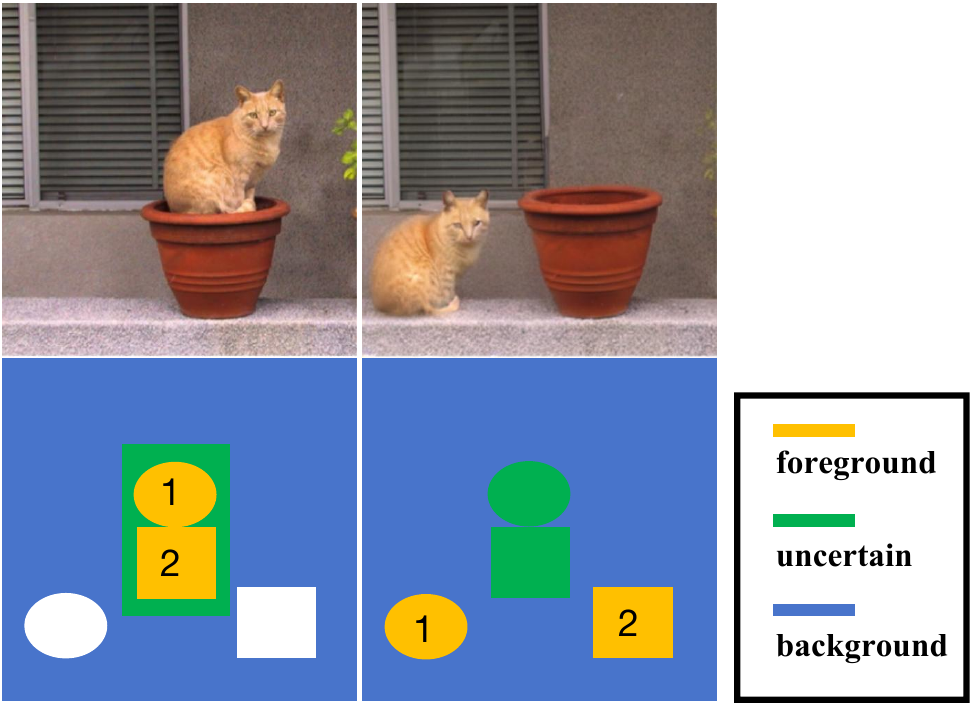}

   \caption{\textbf{Regions division. }
   Left: original image and its region division. Right: edited image and its region division.
   The edited image is segmented into three regions, each marked with a unique color, and the corresponding areas in the original image are highlighted with the same colors }
   \label{fig:regions}
\end{figure}

\noindent Based on these observations, we can make full use of the layout prior information to correct those wrong matches. Specifically, we decompose the edited result image into three regions: the foreground region, the background region and the uncertain region as shown in \cref{fig:regions}.
For the background region, we do not need to perform feature similarity calculation and can directly perform pixel-to-pixel matching. For the foreground region, we perform instance-level feature matching, which is very necessary when there are semantically similar objects in the image. For the uncertain region, since there is no corresponding area in the original image, it actually requires to repair the missing region by using image inpainting. Inspired by the method of Matting \cite{MGMatting}, we obtain a transitional area in the foreground of the original image through dilation and erosion operations and treat it as the target matching area for the uncertain region. The complete workflow can refer to the pseudo-code as shown in Algorithm 1 in the supplementary material. After getting the corrected projection field $P_c$, \cref{eq:warp} is reformulated as:
\begin{equation}
    V^{src\rightarrow tar}_t = W\left(V^{src}_t,P_c\right)
  \label{eq:warp2}
\end{equation}

\subsection{Asynchronous Editing}
\begin{figure}[htbp]
  \centering
   \includegraphics[width=1.0\linewidth]{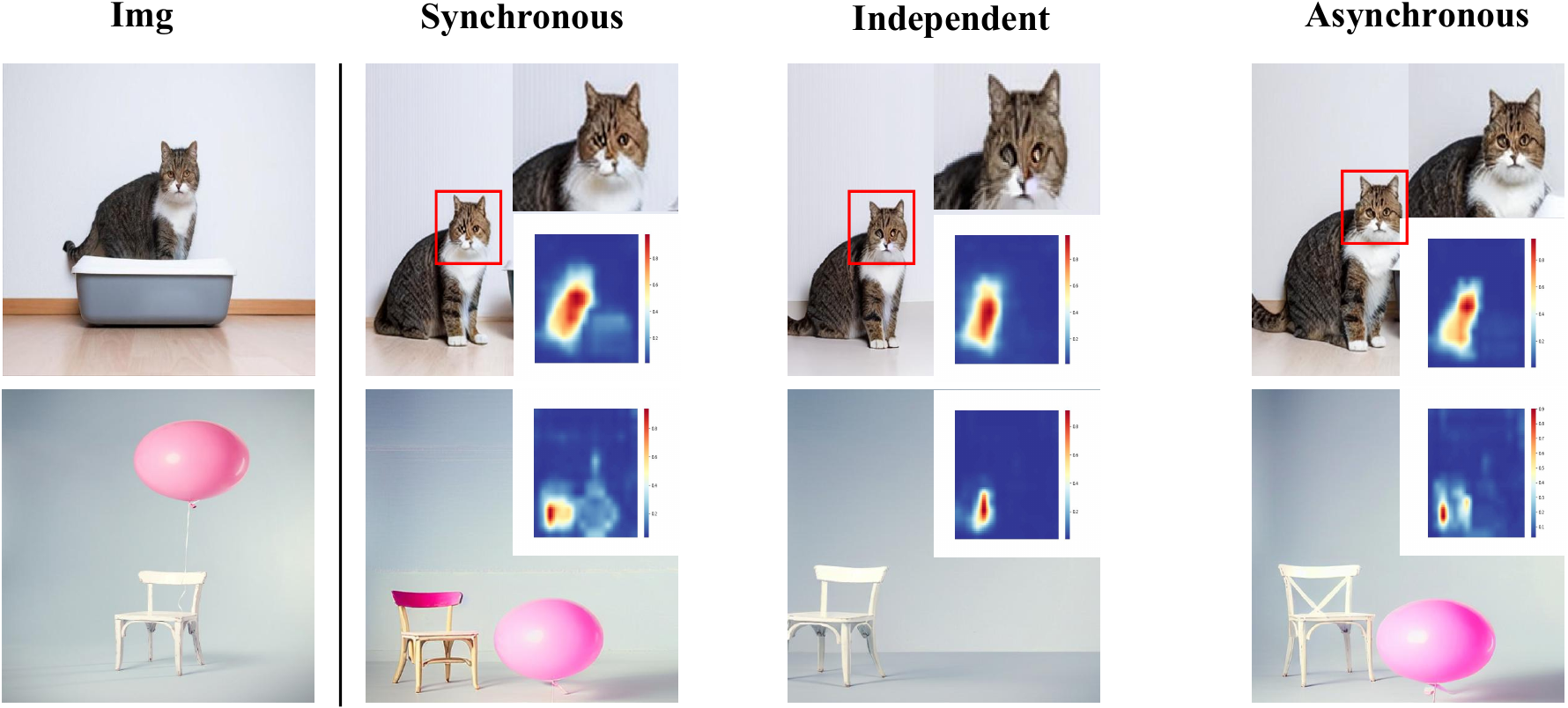}

   \caption{\textbf{Concepts entanglement in synchronous editing. }
   The cross-attention maps and the edited results for synchronous editing, independent editing and asynchronous editing. It shows that editing simultaneously can lead to the entanglement of multiple concepts (in the second column, the cross-attention maps corresponding to "chair" and "cat" show high response for "pot" and "balloon"), which diminishes the appearance fidelity. Asynchronous editing can alleviate this issue.}
   \label{fig:Asyn}
\end{figure}

\begin{figure}[htbp]
  \centering
   \includegraphics[width=1.0\linewidth]{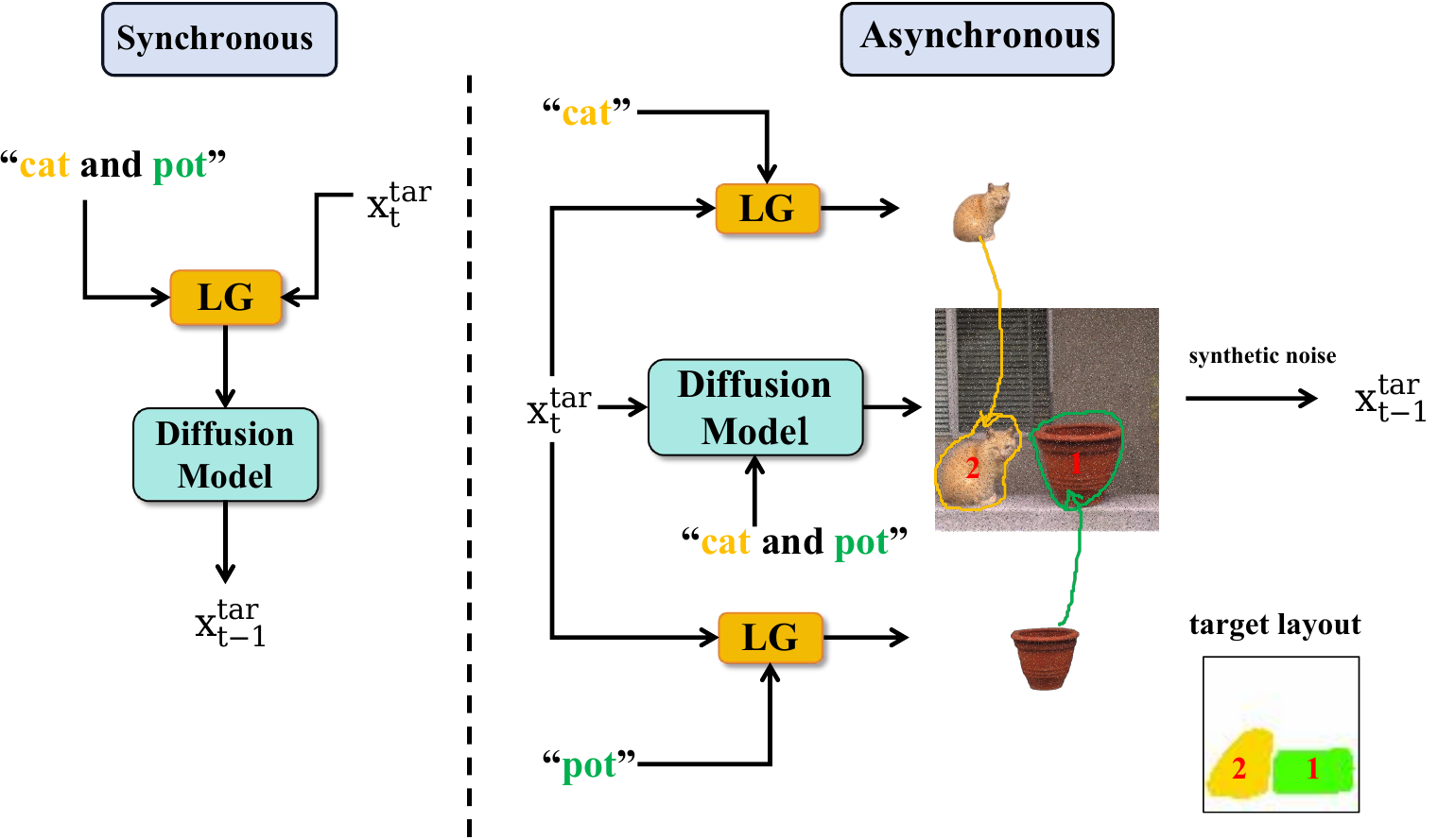}

   \caption{\textbf{Comparison between synchronous pipeline and asynchronous pipeline. }}
   \label{fig:AE-pipeline}
\end{figure}

When performing layout transformation in CLED \cite{zhang2023continuous}, region loss for all objects is calculated by using \cref{eq:mean region loss}, and then its gradient is used to update the latent through a single backward propagation.
However, we find that this synchronous editing method might encourage concepts entanglement, which weakens the fidelity of the objects as shown in \cref{fig:Asyn}. 
When trying to edit the objects independently, this phenomenon disappears and the appearance of objects is kept perfectly as shown in the 3-th column of \cref{fig:Asyn}. Consequently, we consider that it is very necessary to alleviate the concepts entanglement. Inspired by OMG \cite{kong2024omg} and MultiDiffusion \cite{bar-tal2023multidiffusion}, we propose an asynchronous scheme to edit each object independently, followed by fusing the results. Specifically, in each denoising step, we use the region loss to perform layout guidance for each object separately, and then input the optimized latent into the U-Net \cite{ho2020denoising} to obtain the predicted noise. Finally, all of object noise will be fused to obtain the synthetic noise according to the target layout. The specific process is shown in \cref{fig:AE-pipeline}. 
In this way, we can effectively alleviate concepts entanglement, as shown in the 4-th column of \cref{fig:Asyn} .
\subsection{Initial Noise Design For Layout Editing}
\label{subsec:method-LFIN}
Initial noise plays a crucial role in shaping the layout of the images. Layout Guidance \cite{chen2024training} has discovered that initial noise contains intrinsic layout information. We also find that when using different initial noise to perform layout transformation with a fixed model, failure happens once in a while. Therefore, we believe that if the layout information contained in the initial noise is closer to the target layout specified by users, it may be more conducive to performing layout adjustment. Motivated by this idea, we propose a layout-friendly initialization noise (LFIN) strategy. 
\noindent Firstly, we crop and paste the objects from the original image to a blank background based on the target layout, and then perform DDIM \cite{song2020denoising} inversion on this composite image, reversing it until $t = 0.7$. Finally, to ensure the diversity of the editing results, we blend the inverted noise image with random noise. The complete pipeline is provided in supplementary material.


\section{Experiments}
\label{sec:experiments}

\subsection{Experimental Setup}
\textbf{Baselines.} We have implemented our method and made comparisons with several baselines that are designed by ourselves or from existing methods: Image-level manipulation (Image-level), DesignEdit \cite{jia2024designeditmultilayeredlatentdecomposition}, and CLED \cite{zhang2023continuous}.
For the Image-level manipulation, we first apply an inpainting model (e.g., Lama \cite{suvorov2022resolution} and TFill \cite{zheng2022bridging}) to fill the object areas in the original image, and then insert the objects into the desired areas through a crop\&paste way. 
DesignEdit \cite{jia2024designeditmultilayeredlatentdecomposition} simulates Photoshop-like editing by representing each object as an independent layer within the latent space, allowing operations such as translation and scaling before merging these layers to generate the final output. Both methods rely on the manual interaction.
CLED \cite{zhang2023continuous} is the closest method to our method in this paper, directly generating results without manual intervention.

\noindent \textbf{Dataset.} Due to the lack of public dataset for evaluating this task, we collect Layout-Bench that includes 25 high-quality images involving different scenes, and each sample consists of 2-3 objects and 1-3 target layouts. We refer readers to the supplementary material for more details about Layout-Bench.

\noindent \textbf{Metrics.} 
We utilize both objective and subjective metrics to evaluate the experimental results. objective metrics include visual similarity and layout alignment. For the visual similarity, we use CLIP \cite{radford2021learning} and DINO \cite{caron2021emerging} features to assess the consistency respectively. For the layout alignment, we adopt the metric from CLED \cite{zhang2023continuous}. Moreover, we use GPT-4O to evaluate the quality of the edited images. For the subjective evaluation, we also conduct a user study. We randomly invite 45 participants, and each questionnaire consists of 14 questions. Participants are asked to rate the object consistency, layout alignment, image quality of the editing results obtained by different methods. We refer readers to the supplementary material for more details about the metrics and user study.

\subsection{Qualitative Comparison}
\begin{figure}[t]
   \includegraphics[width=1.0\linewidth]{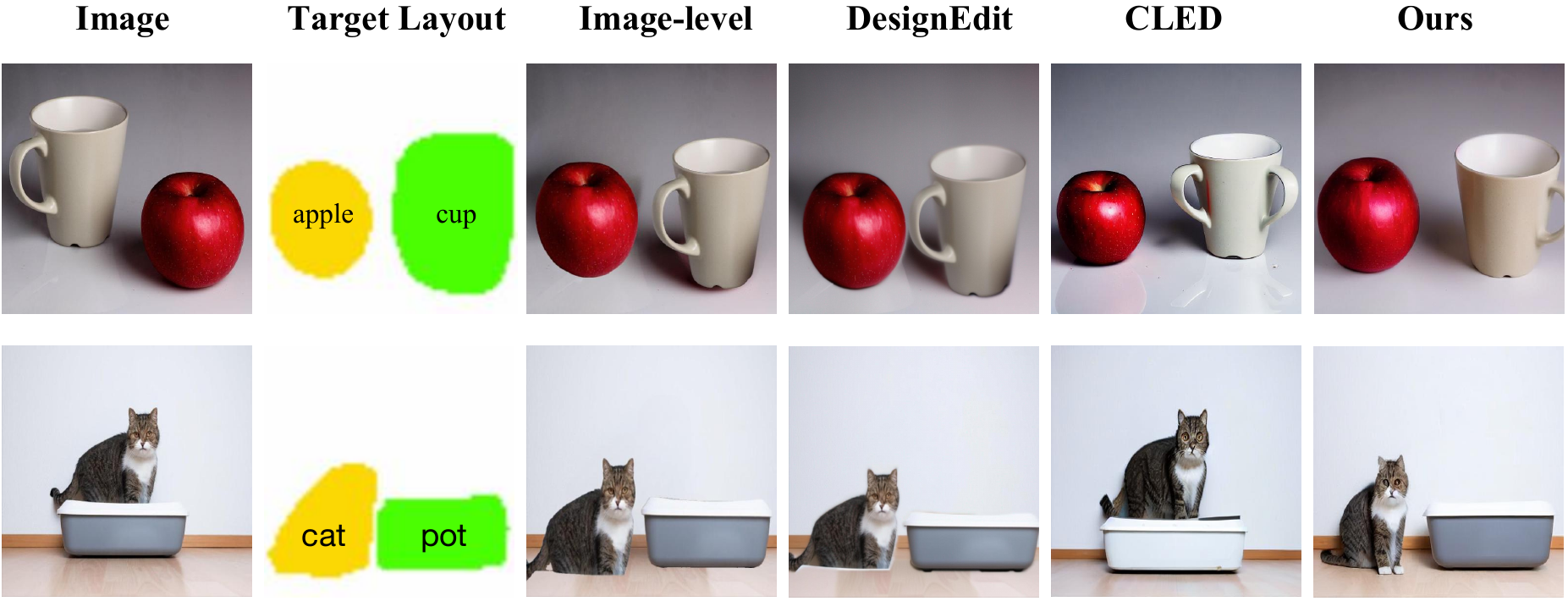}

   \caption{\textbf{Qualitative comparison with baselines.}
   Compared with Image-level manipulation, DesignEdit \cite{jia2024designeditmultilayeredlatentdecomposition} and CLED \cite{zhang2023continuous}. Our method has advantage in handling occlusions and generating realistic lighting and shadow effects.}
   \label{fig:Qualitative-Comparison}
\end{figure}
The qualitative comparison results are shown in \cref{fig:Qualitative-Comparison}. Image-level and DesignEdit \cite{jia2024designeditmultilayeredlatentdecomposition} have similar results. Although DesignEdit \cite{jia2024designeditmultilayeredlatentdecomposition} looks more consistent due to its integration in the latent space, neither of them produces entirely satisfactory results.
Both of them are difficult to ensure that the object is consistent with the surrounding environment and generate realistic lighting effects.
Moreover, when occlusion occurs in the original image, the missing parts of the object cannot be effectively reconstructed as shown in the example of ``cat-pot'', where the cat's feet are occluded by the pot in original image, and the editing result still lacks feet.
Compared to Image-level and DesignEdit \cite{jia2024designeditmultilayeredlatentdecomposition}, CLED \cite{zhang2023continuous} generates more realistic lighting and shadow effects, but it struggles with layout adjustment, and the appearance of the object after editing undergoes significant changes. For example in the ``apple-cup'' case, both of the apple and cup change significantly after editing. In contrast, our method gains significant improvements in the accuracy of layout alignment and the consistency of object appearance. At the same time, our method is capable of generating realistic lighting effects. Besides, when there is occlusion in the original image, our method can automatically repair the missing parts of the target. More results can be found in the supplementary material.

\subsection{Quantitative Comparison}

\noindent The results of the quantitative comparison are shown in \cref{tab:Quantitative-results}. For the visual similarity, our method is slightly lower than the Image-level and DesignEdit methods, as both of them directly paste the objects from the original image into the desired position. Compared to CLED \cite{zhang2023continuous}, our method achieves better visual similarity. In terms of layout alignment, our method significantly outperforms CLED \cite{zhang2023continuous}, and is the same as DesignEdit \cite{jia2024designeditmultilayeredlatentdecomposition}. Considering that DesignEdit \cite{jia2024designeditmultilayeredlatentdecomposition} involves manual intervention, it suggests that our method can improve the accuracy of manual operations in terms of layout alignment. In terms of image quality, the edited images produced by our method outperform all baselines.
\begin{table}[htbp]
  \centering
  \tabcolsep=0.1cm
  \begin{tabular}{lclclclc}
    \toprule
      & \multicolumn{2}{c}{\makecell[c]{Visual\\similarity} $\left(\uparrow\right)$} & \makecell[c]{Layout\\alignment}$\left(\uparrow\right)$ & \makecell[c]{Image\\quality}$\left(\uparrow\right)$  \\
    \cmidrule(lr){2-3}
     & DINO & CLIP &\\
    \midrule
    Image-level & \textbf{0.905} & \textbf{0.955} & \textbf{0.023} & 7.442 \\
    DesignEdit & 0.898 & 0.949 & 0.021 & 7.115 \\
    CLED & 0.849 & 0.944 & 0.015 & 8.143 \\
    Ours & 0.856 & 0.942 & 0.021 & \textbf{8.501} \\
    \bottomrule
  \end{tabular}
  \caption{\textbf{Quantitative comparison with baselines.} }
  \label{tab:Quantitative-results}
\end{table}

\begin{figure}[htbp]
   \includegraphics[width=1.0\linewidth]{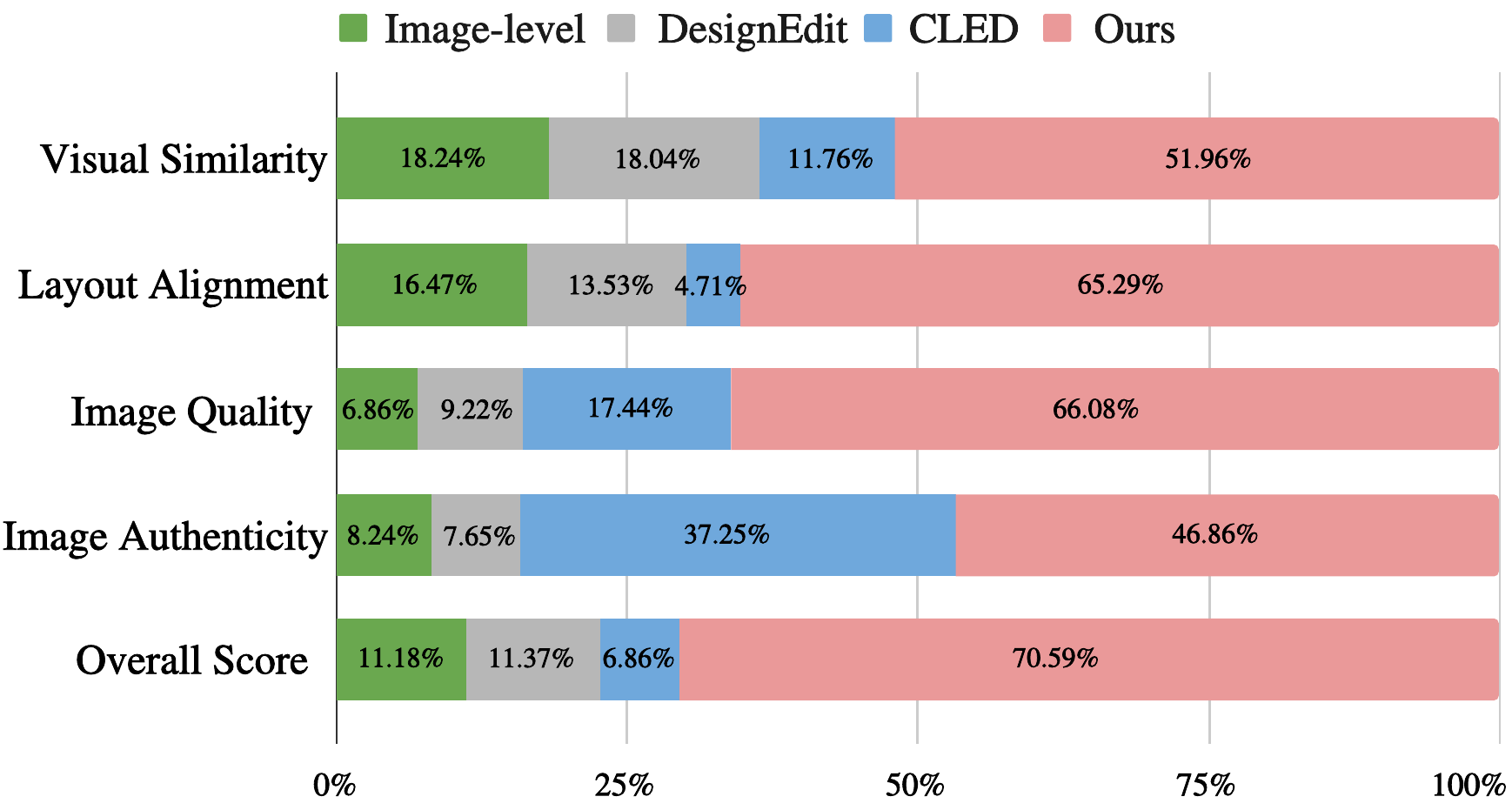}

   \caption{\textbf{User study} 
   }
   \label{fig:user-study}
\end{figure}
\noindent \textbf{User Study.} \cref{fig:user-study} shows the user study results, where our method significantly surpasses all baselines with respect to most metrics.

\subsection{Ablation Study}\label{sec:abalation}
To assess the effectiveness of various components involved in our framework, we conduct an ablation study encompassing the following elements: Multi-Concept Learning (MCL), Region Prior Appearance Projection (RPAP), Asynchronous Editing (AE) and  Layout-Friendly Initialization Noise (LFIN).

\begin{figure}[htbp]
   \includegraphics[width=1.0\linewidth]{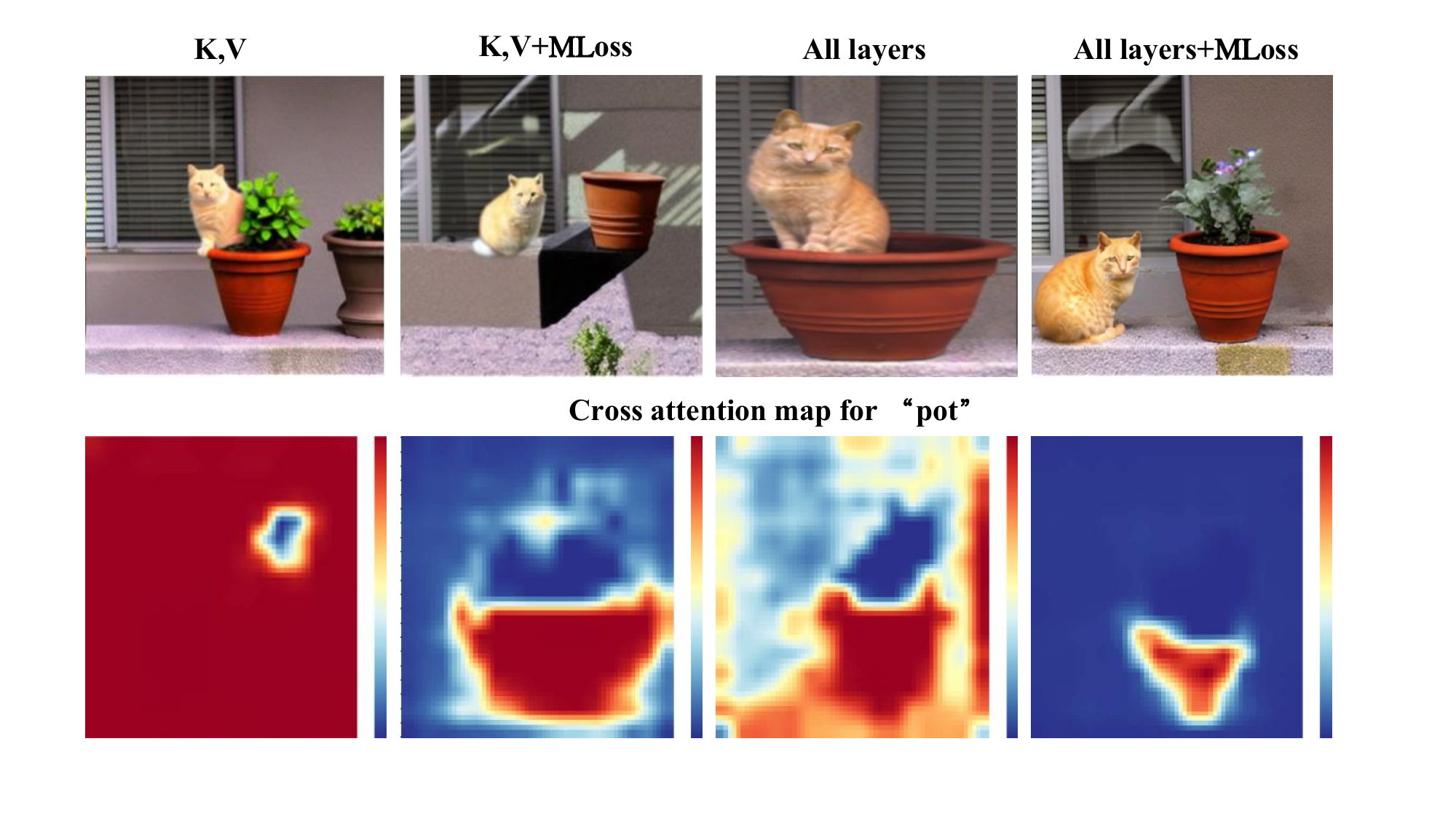}

   \caption{\textbf{Ablation study on masked diffusion loss. }
   Top: Editing results by fine-tuning different layers with and without masked diffusion loss. Bottom: Cross-attention map correspond to the ``pot'' word. Attention regions are spread across the entire image when masked diffusion loss is not applied, and layout transformation fails.
   }
   \label{fig:ablation-mloss}
\end{figure}

\begin{figure}[htbp]
   \includegraphics[width=1\linewidth]{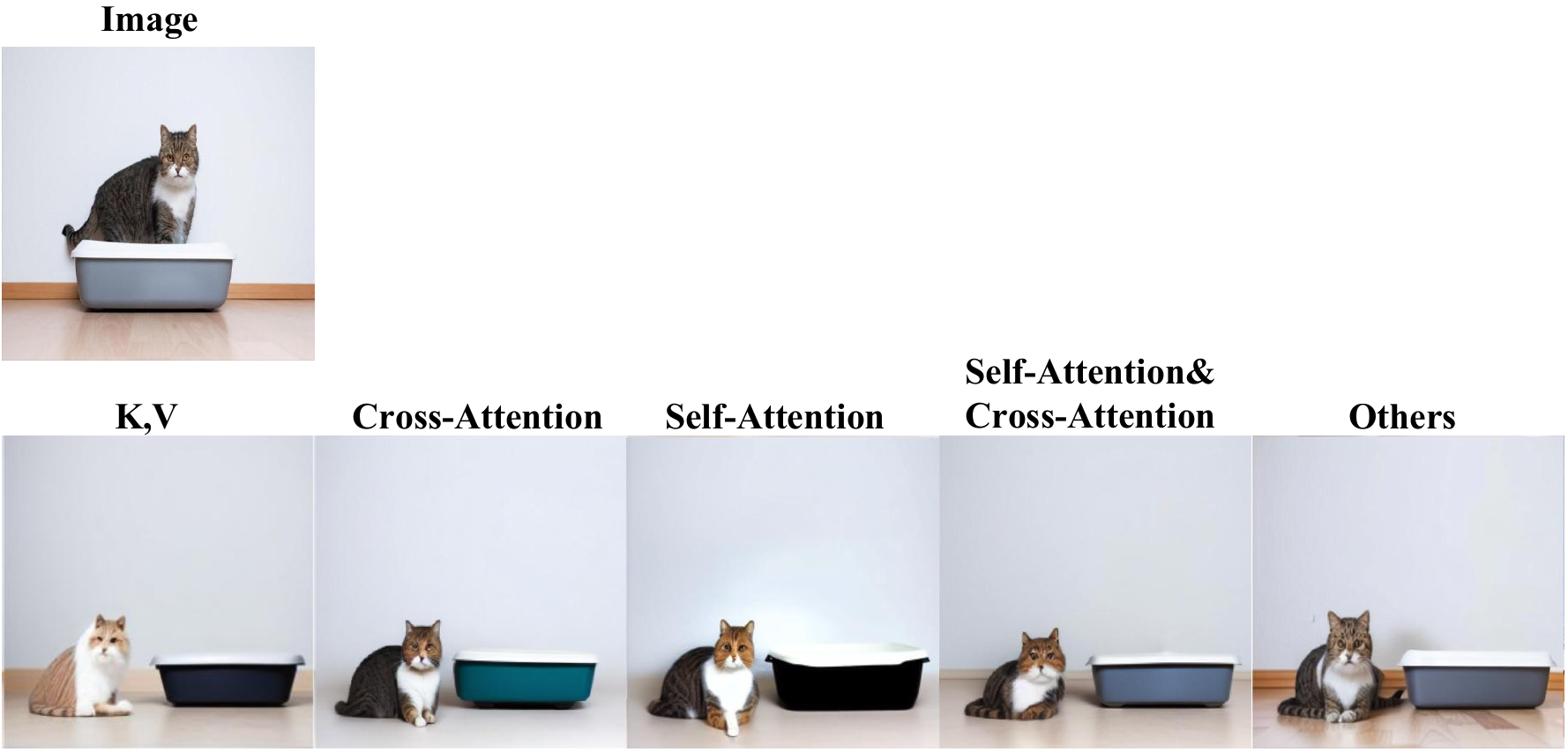}
   \caption{\textbf{Ablation study on fine-tuning layers. }
   Fine-tune different layers in diffusion models with masked diffusion loss. Because of masked diffusion loss, all fine-tuning schemes reshape the layout successfully.}
   \label{fig:finetune}
\end{figure}

\noindent\textbf{MCL.} We conduct ablation experiments on the multi-concept learning method introduced in \cref{subsec: mcl}, which focuses on validating the effectiveness of the model fine-tuning method in the second training stage. Firstly, we fine-tune the key (K) and value (V) weights of the cross-attention modules, as well as all layers in the U-Net \cite{song2020denoising}, both with and without masked diffusion loss. As shown in \cref{fig:ablation-mloss}, without the use of masked diffusion loss, layout transformation fails regardless of whether K,V or all layers are fine-tuned. When examining the cross-attention map, we observe that the attention regions are dispersed across the entire image. The experimental results above demonstrate the important role of masked diffusion loss in the fine-tuning stage.
At the same time, we find that layout transformation is successful when fine-tuning K,V layers with masked diffusion loss, but noticeable changes in the appearance exits. So we apply masked diffusion loss to fine-tune different layers within the diffusion models, as illustrated in \cref{fig:finetune}. All fine-tuning schemes are able to achieve layout transformation successfully, further proving the important role of masked diffusion loss during the fine-tuning stage. 
For most of fine-tune schemes in \cref{fig:finetune}, however, the appearance of the objects changes significantly. Interestingly, in contrast to the conclusion in CD \cite{kumari2023CD}, where only the K, V layers are considered as the most important layers, fine-tuning these layers results in the most noticeable changes of the objects appearance. However, when fine-tuning the others layers with less significance, it maintains better appearance properties.

\begin{figure}[t]
   \includegraphics[width=1.0\linewidth]{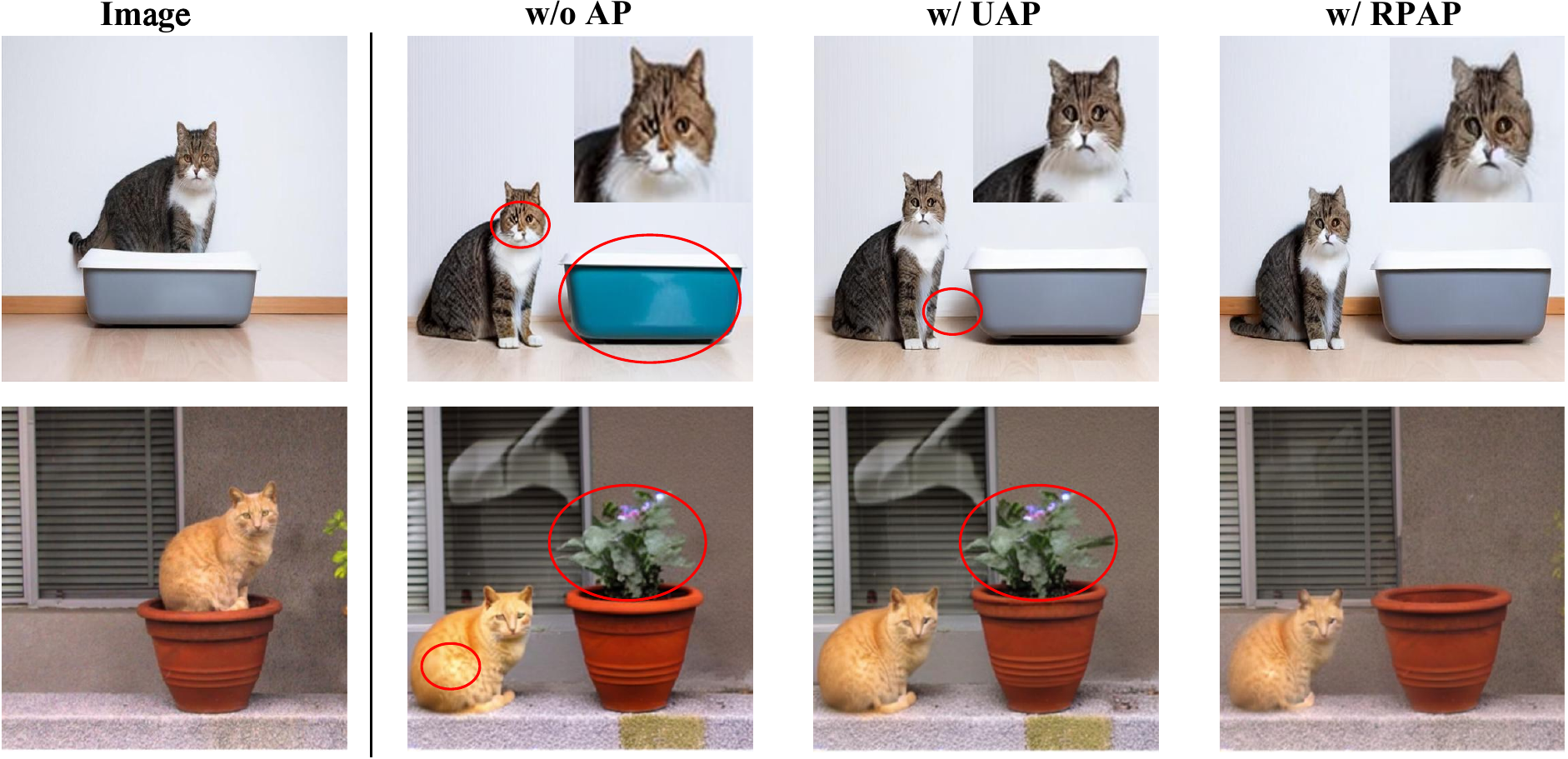}

   \caption{\textbf{Ablation study on UAP\&RPAP. }}
   \label{fig:PRAP-exp}
\end{figure}

\noindent\textbf{UAP\&RPAP.} \cref{fig:PRAP-exp} shows the ablation results of UAP \& RPAP. 
When appearance projection is not applied, the appearance of the objects changes significantly after editing, which is due to the layout guidance methods based on latent optimization (e.g., the cat's face is distorted as shown in the second column). Appearance of the objects can be effectively repaired with UAP (as shown in the third column in \cref{fig:PRAP-exp}, where the cat's face and the color of pot is repaired after projection). However, due to the existence of incorrect matching, it can result in incorrect appearance mapping (e.g., the plants growing out of the flowerpot are retained). RPAP can effectively correct the incorrectly matched feature points and improve the fidelity of the targets.

\begin{figure}[t]
   \includegraphics[width=1.0\linewidth]{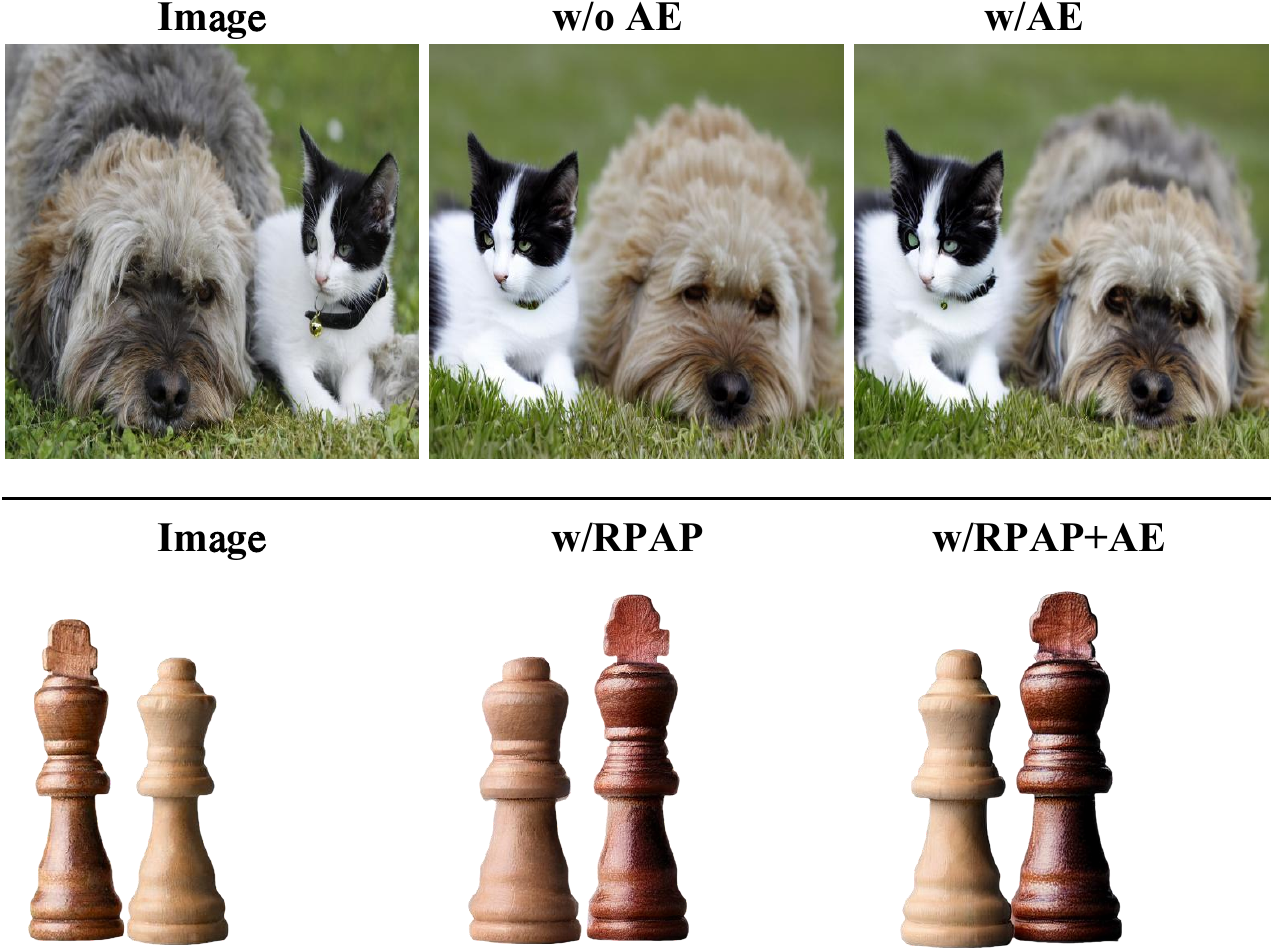}
   \caption{\textbf{Ablation study on AE. } The first row compares Synchronous Editing (w/o AE) with Asynchronous Editing (w/AE). The second row shows that even with the use of appearance projection in some cases, the appearance still cannot be perfectly restored because of concept entanglement.}
   \label{fig:AONF-exp}
\end{figure}

\noindent\textbf{AE.} \cref{fig:AONF-exp} shows the qualitative ablation results of AE. Even with the use of RPAP in some cases, it is still unable to completely maintain the appearance characteristics as shown in the bottom of \cref{fig:AONF-exp}. When AE is applied, it can further improve the fidelity of the objects.

\begin{figure}[t]
   \includegraphics[width=1.0\linewidth]{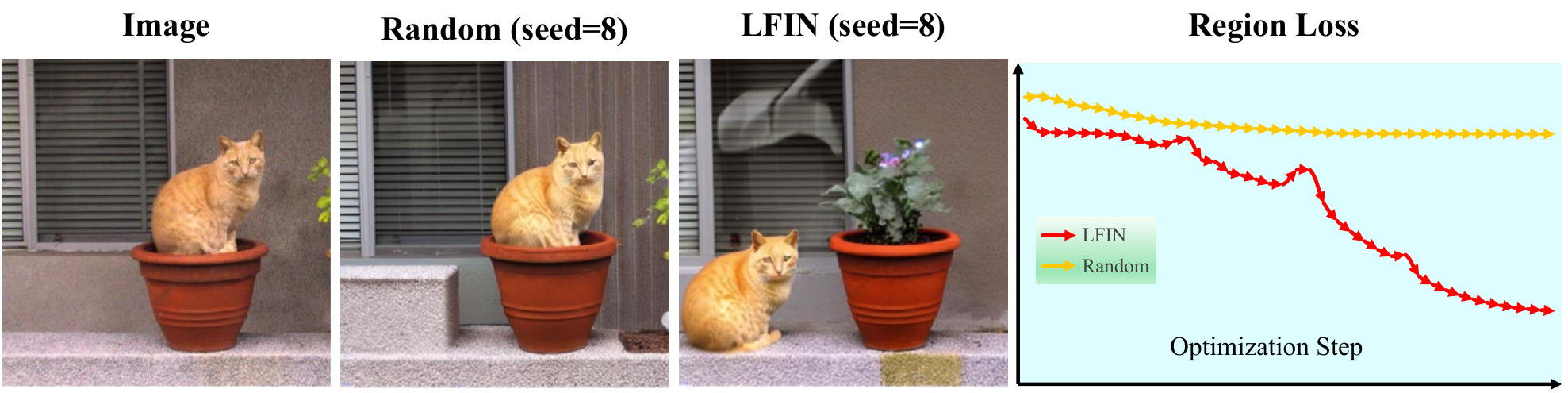}
   \caption{\textbf{Ablation study on initial noise. }
   }
   \label{fig:LFIN-exp}
\end{figure}

\noindent\textbf{LFIN.} 
\cref{fig:LFIN-exp} shows the ablation results of LFIN. The two middle images display the editing results by using random noise and LFIN, respectively. The right image shows the curve of the region loss when using different initial noise. It can be seen that LFIN noise successfully changes the layout of the original image, while the random initial noise fails. At the same time, the convergence speed of the region loss is significantly improved after using LFIN as the initialization noise, indicating that LFIN is more effective in facilitating the layout editing.
\section{Conclusion}
\label{sec:conclu}

This paper presents a two-stage image layout editing method, which utilizes the natural semantic consistency explored in this task. By employing layout guidance followed by appearance projection, our method can ensure the consistency of the objects after layout re-arrangement. Extensive experiments show that our method outperforms previous works in both layout alignment accuracy and image quality after the layout editing.

As the further work, we plan to improve the computational efficiency of the proposed method and handle more complex scenarios such as severe occlusions. Besides, it is also promising to extend our method to edit the layout of three-dimensional scenes.
{
    \small
    \bibliographystyle{ieeenat_fullname}
    \bibliography{main}
}


\end{document}